\documentclass[conference]{IEEEtran}

\usepackage{cite}
\usepackage{amsmath,amssymb,amsfonts}
\usepackage{graphicx}
\usepackage{textcomp}
\usepackage{booktabs}
\usepackage{url}
\usepackage{multirow}

\def\BibTeX{{\rm B\kern-.05em{\sc i\kern-.025em b}\kern-.08em
    T\kern-.1667em\lower.7ex\hbox{E}\kern-.125emX}}

\begin{document}

\title{Rethinking Closed-Ended Medical VLM Evaluation: From Stable Wrongness to Counterfactual Answer Sensitivity}

\author{
\IEEEauthorblockN{
Qi Wu\textsuperscript{2,*},
Xiaoyu Liu\textsuperscript{1,4,*},
Ao Wang\textsuperscript{1,4},
Rongsheng Wang\textsuperscript{2},
Wenting Chen\textsuperscript{3},
Wenxuan Wang\textsuperscript{1,4},
Qingsong Yao\textsuperscript{1,4,\dag}
}

\IEEEauthorblockA{
\textsuperscript{1}Renmin University of China\quad
\textsuperscript{2}University of Science and Technology of China\quad
\textsuperscript{3}Stanford University\quad\\
\textsuperscript{4}Beijing Key Laboratory of Smart Healthcare and AI Driven Health Governance\quad\\
\textsuperscript{*}These authors contributed equally.\quad
\textsuperscript{\dag}Corresponding author: Qingsong Yao (qingsongyao98@gmail.com)
}
}

\maketitle

\begin{abstract}
As medical VLMs are increasingly evaluated for clinical decision-support scenarios, knowing when to trust their predictions is critical. Closed-ended medical VQA has become a common benchmark format because it enables automated scoring. Yet, systematic investigation into whether this format changes the reliability problem remains scarce. We identify a failure mode that we call \emph{stable wrongness}: models can produce incorrect yes/no answers that remain confident and repeatable even when patient-specific visual evidence is removed or mismatched. This failure violates the premise behind sampling-stability uncertainty methods, including recent hallucination-aware calibration (HAC) variants: disagreement is informative only when errors are unstable. When shortcut-driven errors are themselves stable, sampling agreement can remain deceptively high. Motivated by this limitation, we introduce Counterfactual Answer Sensitivity (CAS), which shifts reliability estimation from answer stability to evidence sensitivity. CAS uses cross-modal counterfactual interventions, such as text-only, black, and corrupted images, to audit whether a committed yes/no answer remains supported when visual evidence is invalidated. Across 7 VLMs and 3 benchmarks, CAS outperforms the best HAC-family scores across all evaluated settings, achieving a macro-averaged AUROC gain of +0.150 for error detection. Mechanistic analysis further shows that CAS recovers hidden errors in the HAC-confident region and that its strongest standalone signal is counterfactual decision-boundary shift rather than hard prediction stability. Our findings suggest that closed-ended evaluation should test visual dependence, not only answer stability, in safety-critical medical contexts.
\end{abstract}

\section{Introduction}
Vision-language models (VLMs) are increasingly evaluated for clinical decision support, making reliable uncertainty estimation essential~\cite{li2023llavamed,tu2024medpalm,zhu2025trust}. Closed-ended medical VQA is attractive because yes/no answers permit inexpensive automated scoring, but reliable clinical reasoning requires predictions to depend on patient-specific visual evidence rather than language habits, dataset priors, or shortcut cues~\cite{agrawal2018dont,geirhos2020shortcut,zech2018variable,degrave2021ai}.

Recent reliability methods emphasize multi-sample agreement, including semantic entropy and hallucination-aware calibration (HAC)~\cite{kuhn2023semantic,farquhar2024detecting,liao2025vase,byun2026overconfidence}. These methods are informative when errors are unstable. We show a complementary failure mode, \emph{stable wrongness}: an incorrect yes/no answer can remain confident and repeatable even when the image is removed, blacked out, or mismatched. A shortcut can therefore be as stable as genuine visual evidence, making sampling agreement ambiguous.

We introduce Counterfactual Answer Sensitivity (CAS), which asks whether the committed answer is sensitive to patient-specific visual evidence. CAS measures target-support and opposite-answer boundary shifts under text-only, black, noisy, and blurred visual counterfactuals. Across 7 VLMs and 3 benchmarks, CAS improves macro-average error-detection AUROC by 0.150 over the best HAC-family score. The code is available at \url{https://github.com/Wanye1307/CAS}.

Our contributions are summarized as follows:
\begin{itemize}
    \item We identify stable wrongness as an important reliability failure in closed-ended medical VQA, showing that models can preserve confident yes/no errors under invalid or mismatched visual evidence.
    \item We introduce closed-ended CAS, an answer-conditioned evidence-sensitivity framework that maps target-support shifts, opposite-answer boundary shifts, and prediction persistence under controlled visual counterfactuals.
    \item We benchmark 7 models across 3 datasets and show that CAS consistently improves over HAC-family sampling-stability pipelines, especially by recovering hidden errors in HAC-confident regions.
\end{itemize}

\section{Related Work}
\textbf{Medical VLM evaluation.} Medical VQA benchmarks such as VQA-RAD, SLAKE, and VQA-Med-2019 contain closed-ended questions that enable exact scoring~\cite{lau2018vqarad,liu2021slake,benabacha2019vqamed}. Broader evaluations increasingly consider factuality, uncertainty, safety, and robustness~\cite{li2023llavamed,tu2024medpalm,zhu2025trust}, yet closed-ended scoring can obscure whether a correct-looking answer is visually grounded.

\textbf{Confidence and sampling uncertainty.} Reliability scores use likelihood, entropy, margins, calibration, or repeated-sample agreement~\cite{guo2017calibration,nixon2019measuring,wang2023selfconsistency,kuhn2023semantic,farquhar2024detecting}. HAC-style medical VQA methods augment confidence with hallucination or grounding signals~\cite{liao2025vase,byun2026overconfidence}. Their blind spot is systematic error: if a shortcut repeatedly produces the same wrong answer, agreement remains high.

\textbf{Counterfactual visual evidence.} Image removal, masking, corruption, and replacement can reveal non-visual priors and shortcut dependence~\cite{goyal2017making,goyal2019counterfactual,ross2017right}. CAS adapts this principle to closed-ended medical VQA by conditioning on the committed answer and measuring structured support and boundary shifts across controlled visual interventions.

\section{Rethinking: When Stability Does Not Imply Visual Grounding}
\label{sec:rethinking}

Closed-ended medical VQA is often treated as the simpler and more controlled evaluation setting because the model only needs to choose between \textit{yes} and \textit{no}. However, this format also changes the reliability problem. A binary answer space compresses the model's decision into two reusable answer labels, making it possible for language priors, prompt templates, and input-format cues to dominate the final decision. The resulting failure is not necessarily noisy or volatile. A closed-ended prediction can be wrong, confident, and highly repeatable at the same time.

This section audits whether such stability reflects visual grounding or stable wrongness. We focus on the central diagnostic: counterfactual visual invalidation. By preserving the question while removing, blacking out, or mismatching the image, we test whether the paired medical image is a necessary condition for the model's yes/no answer. This diagnostic identifies an evidence-insensitive failure mode and motivates reliability estimation based on counterfactual visual evidence sensitivity.

\subsection{Invalid visual evidence preserves confident answers}

We test whether the yes/no prediction is grounded in the provided image. For each closed-ended question, we keep the text fixed and replace the visual input with one of three counterfactual conditions: no image (\textit{text-only}), a homogeneous black image, or a randomly mismatched medical image from the same benchmark. The real image serves as the baseline. This diagnostic does not train a detector; it directly measures how the answer distribution changes when patient-specific visual evidence is removed or mismatched.

For each input condition $c$, we report binary accuracy, mean binary confidence, yes rate, absolute yes-rate shift relative to the real-image condition, and agreement with the real-image prediction:
\begin{equation}
\mathrm{Conf}^{(c)}
=
\max\left(p^{(c)}_{\mathrm{yes}}, p^{(c)}_{\mathrm{no}}\right),
\end{equation}
\begin{equation}
\Delta_{\mathrm{yes}}^{(c)}
=
\left|
\Pr\left(\hat{y}^{(c)}=\mathrm{yes}\right)
-
\Pr\left(\hat{y}^{(\mathrm{real})}=\mathrm{yes}\right)
\right|,
\end{equation}
\begin{equation}
\mathrm{Agree}^{(c)}
=
\Pr\left(\hat{y}^{(c)}=\hat{y}^{(\mathrm{real})}\right).
\end{equation}

\begin{table}[t]
\centering
\caption{Closed-ended counterfactual image probe averaged over 21 model--dataset settings. Counterfactual inputs preserve the question but invalidate patient-specific visual evidence. $\Delta$Yes denotes the absolute yes-rate shift relative to the real-image condition; Agree is agreement with the real-image prediction.}
\label{tab:counterfactual_image_probe}
\small
\setlength{\tabcolsep}{2.7pt}
\begin{tabular}{@{}lccccc@{}}
\toprule
\textbf{Input} & \textbf{Acc.} & \textbf{Conf.} & \textbf{Yes} & $\boldsymbol{\Delta}$\textbf{Yes} & \textbf{Agree} \\
\midrule
Real image & 0.762 & 0.871 & 0.487 & 0.000 & 1.000 \\
Text only & 0.544 & 0.790 & 0.496 & 0.131 & 0.604 \\
Black image & 0.540 & 0.794 & 0.304 & 0.214 & 0.573 \\
Random image & 0.560 & 0.901 & 0.294 & 0.193 & 0.619 \\
\bottomrule
\end{tabular}
\end{table}

Table~\ref{tab:counterfactual_image_probe} reveals two complementary effects. First, invalidating the image does not make the models uncertain. Text-only and black-image inputs still produce high binary confidence, around $0.79$, while random images produce even higher confidence, $0.901$ on average. Second, the visual input state systematically shifts the yes/no prior. Compared with the real-image yes rate of $0.487$, black images and random images reduce the yes rate to $0.304$ and $0.294$, respectively. These shifts indicate that input state can substantially affect the yes/no prior.

At the same time, counterfactual predictions remain substantially correlated with real-image predictions. Text-only, black-image, and random-image inputs agree with the real-image prediction in $60.4\%$, $57.3\%$, and $61.9\%$ of model--sample pairs, respectively. Thus, the model is not simply failing gracefully under missing evidence. It often remains confident and preserves a large fraction of its original yes/no decisions even when the real medical image is absent or mismatched. The paired image is therefore not always a necessary condition for the model's answer.

\subsection{Implication: stability must be tested against evidence sensitivity}

The diagnostics above identify a failure mode that is easy to miss if reliability is measured only by answer stability or raw confidence, a limitation already highlighted in broader uncertainty estimation under dataset shift~\cite{hendrycks2017baseline,ovadia2019uncertainty}. Directly, the counterfactual image probe shows that invalid visual inputs can still produce confident yes/no decisions.

This challenges sampling-majority reliability frameworks such as HAC and semantic-entropy-style estimators~\cite{wang2023selfconsistency,kuhn2023semantic,farquhar2024detecting,liao2025vase,byun2026overconfidence}. These methods are most effective when errors are stochastic: risky answers produce diverse sampled generations and low agreement. However, if an error is caused by a stable answer-label shortcut or an input-format prior, repeated sampling can reproduce the same incorrect answer. In that case, high agreement is not evidence of visual grounding; it is evidence of shortcut consistency.

Therefore, closed-ended reliability estimation should not only ask whether an answer is stable. It should ask whether the committed answer remains supported when valid visual evidence is removed, corrupted, or otherwise made unreliable. CAS turns this diagnostic principle into an error-detection score by measuring target-answer support and yes/no boundary movement under controlled counterfactual visual inputs.

The random-image condition in this section is used as a diagnostic probe for visual-evidence mismatch: it tests whether a model can reproduce the same answer when the patient-specific image is replaced by an unrelated medical image. CAS uses a fixed operational intervention space consisting of text-only, black-image, noisy-image, and blurred-image conditions. These operational interventions do not need to duplicate every diagnostic probe; they quantify the same evidence-sensitivity principle using controlled and reproducible visual counterfactuals.

\section{Counterfactual Answer Sensitivity}
\begin{figure*}
    \centering
    \includegraphics[width=1\linewidth]{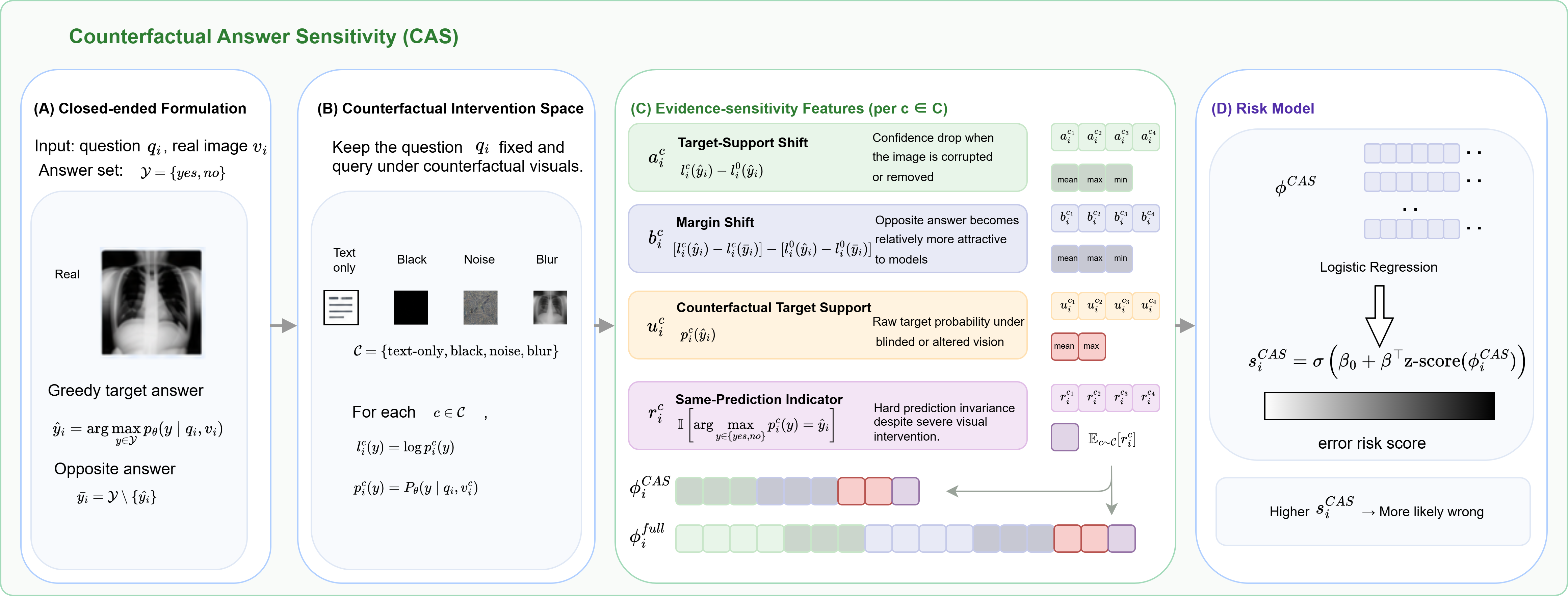}
    \vspace{-3mm}
    \caption{Overview of our proposed counterfactual answer sensitivity (CAS) framework for uncertainty quantification.}
    \vspace{-5mm}
    \label{fig:placeholder}
\end{figure*}
Closed-ended medical VLM predictions can remain confident under missing or mismatched visual evidence. These failures are difficult to capture with confidence alone: a model can be highly confident both on the real image and on counterfactual inputs that contain no valid medical evidence. The key missing quantity is therefore not only how confident the model is, but whether the predicted answer is specifically supported by the real image.

This observation motivates our Counterfactual Answer Sensitivity (CAS) evaluation, following the counterfactual evidence perspective~\cite{goyal2019counterfactual}. Rather than treating the real-image confidence as a sufficient uncertainty signal, CAS explicitly compares the support for the same predicted answer under the real image and under controlled counterfactual inputs such as text-only, black, noise, and blurred images. If an answer remains strongly supported after the image is removed or corrupted, CAS assigns higher risk; if the real image uniquely strengthens the predicted answer, CAS assigns lower risk. In this way, CAS operationalizes the failure patterns identified above: it penalizes answers that are stable under missing or mismatched visual evidence and rewards answers whose support is tied to the real medical image.

\subsection{Closed-ended formulation}

Consider a sample $i$ with question $q_i$, real image $v_i$, and a binary answer set $\mathcal{Y}=\{\mathrm{yes},\mathrm{no}\}$. The model first produces a deterministic target answer:
\begin{equation}
    \hat{y}_i = \arg\max_{y\in\mathcal{Y}} p_\theta(y \mid q_i, v_i),
\end{equation}
and the opposite alternative answer is denoted by:
\begin{equation}
    \bar{y}_i =
    \begin{cases}
    \mathrm{no}, & \hat{y}_i=\mathrm{yes},\\
    \mathrm{yes}, & \hat{y}_i=\mathrm{no}.
    \end{cases}
\end{equation}
We establish a parallel intervention space by querying the model with the same textual prompt $q_i$ across a set of controlled counterfactual visual conditions:
\begin{equation}
    \mathcal{C}=\{\text{text-only},\text{black},\text{noise},\text{blur}\}.
\end{equation}
For the original pristine state, we denote its log probability as $l_i^0(y) = \log P_{\theta}(y \mid q_i, v_i)$. For each counterfactual condition $c \in \mathcal{C}$ with altered visual input $v_i^c$, the decoder's binary probability $p_i^c(y) = P_{\theta}(y \mid q_i, v_i^c)$ and its corresponding log probability $l_i^c(y) = \log p_i^c(y)$ are recorded.

\subsection{Evidence-sensitivity features}

To understand how the model arrives at its decision, CAS extracts four complementary types of signals for each intervention $c$. These features quantify how the model's answer support changes when valid visual evidence is removed or corrupted:

\begin{itemize}
    \item \textbf{Target-Support Shift ($a_i^c$)} This feature measures how much the model's confidence in its original answer drops after we corrupt or remove the image:
    \begin{equation}
        a_i^c = l_i^c(\hat{y}_i) - l_i^0(\hat{y}_i)
    \end{equation}
    If the prediction depends on the image, removing or corrupting the image should reduce support for the committed answer ($a_i^c$ becomes negative). Conversely, if $a_i^c$ stays close to zero or becomes positive, the answer may be supported by non-visual cues.
    
    \item \textbf{Margin Shift ($b_i^c$)} This feature tracks how removing the image changes the relative margin between the committed answer and the opposite answer:
    \begin{equation}
        b_i^c = [l_i^0(\bar{y}_i) - l_i^c(\bar{y}_i)] - [l_i^0(\hat{y}_i) - l_i^c(\hat{y}_i)]
    \end{equation}
    It evaluates whether the counterfactual intervention changes the relative support of the alternative choice. Large shifts indicate that the original yes/no preference is sensitive to the visual intervention.
    
    \item \textbf{Direct Counterfactual Target Support ($u_i^c$)} This is the raw probability the model assigns to its original answer under the altered visual state:
    \begin{equation}
        u_i^c = p_i^c(\hat{y}_i)
    \end{equation}
    It acts as a direct probe of counterfactual answer support. For instance, if we feed the model a completely black image ($c = \text{black}$) and it still outputs a $99\%$ probability for $\hat{y}_i$, this score indicates that the answer remains strongly supported by non-visual cues.
    
    \item \textbf{Same-Prediction Indicator ($r_i^c$)} A binary indicator capturing whether the model preserves its hard prediction choice under a counterfactual visual condition:
    \begin{equation}
        r_i^c = \mathbb{I}\left[\arg\max_{y \in \{\mathrm{yes}, \mathrm{no}\}} p_i^c(y) = \hat{y}_i\right]
    \end{equation}
    This acts as a hard-decision stability test. If the final binary answer (``yes'' vs. ``no'') remains unaffected after the image is blurred or removed, the prediction may not be sufficiently tied to the visual modality.
\end{itemize}

To provide the downstream detector with both detailed condition-specific responses and broad behavioral trends, we organize these features into ordered coordinate vectors representing each signal channel: $\mathbf{a}_i = [a_i^{c_1}, \dots, a_i^{c_4}]^\top \in \mathbb{R}^4$, $\mathbf{b}_i = [b_i^{c_1}, \dots, b_i^{c_4}]^\top \in \mathbb{R}^4$, and $\mathbf{u}_i = [u_i^{c_1}, \dots, u_i^{c_4}]^\top \in \mathbb{R}^4$. 

We then apply a full macro-aggregation operator $\mathcal{T}_{\text{full}}$ and a reduced operator $\mathcal{T}_{\text{sub}}$ to summarize the model's overall response profile across all interventions:
\begin{align}
    \mathcal{T}_{\text{full}}(\mathbf{x}) &= \left[ \operatorname{mean}(\mathbf{x}), \operatorname{max}(\mathbf{x}), \operatorname{min}(\mathbf{x}) \right], \\
    \mathcal{T}_{\text{sub}}(\mathbf{x}) &= \left[ \operatorname{mean}(\mathbf{x}), \operatorname{max}(\mathbf{x}) \right]
\end{align}

Our main benchmark uses a compact, pre-specified 9-dimensional CAS representation that keeps the aggregated behavior of each signal family:
\begin{equation}
    \phi_i^{CAS} = \left[ \mathcal{T}_{\text{full}}(\mathbf{a}_i) \parallel \mathcal{T}_{\text{full}}(\mathbf{b}_i) \parallel \mathcal{T}_{\text{sub}}(\mathbf{u}_i) \parallel \mathbb{E}_{c\sim\mathcal{C}}[r_i^c] \right]^\top \in \mathbb{R}^{9}.
\end{equation}
where $\parallel$ denotes vector concatenation, and $\mathbb{E}_{c\sim\mathcal{C}}[r_i^c]$ represents the same-prediction rate across visual interventions. We also evaluate a full 17-dimensional variant in the ablation study:
\begin{equation}
    \phi_i^{full} = \left[ \mathbf{a}_i^\top \parallel \mathbf{b}_i^\top \parallel \phi_i^{CAS} \right]^\top \in \mathbb{R}^{17}.
\end{equation}

This multi-channel framing allows CAS to capture distinct evidence-sensitivity patterns: mean features evaluate average persistence under interventions, max configurations isolate strong shortcut triggers, and min features test whether evidence sensitivity appears in at least one condition. The compact 9-dimensional representation is used as the default CAS score because it matches the hypothesized signal families and avoids relying on condition-specific feature count as the explanation for performance.

\subsection{Risk model and scope}

To deploy these features for error detection, $\phi_i^{CAS}$ is mapped to a unified error-risk score $s_i^{CAS}$ via a cross-validated, class-balanced $L_2$-regularized logistic regression model:
\begin{equation}
    s_i^{CAS} = \sigma\left(\beta_0 + \beta^\top \operatorname{z-score}(\phi_i^{CAS})\right)
\end{equation}
where $\sigma(\cdot)$ is the standard sigmoid mapping, and $\operatorname{z-score}(\cdot)$ denotes feature normalization using the mean and variance statistics derived from the training fold. The risk model is evaluated out of fold using five-fold stratified cross-validation, which prevents train-test leakage in the reported error-detection scores.

\section{Main Benchmark Results}

We evaluate CAS against all standard confidence baselines and sampling-stability metrics across 7 general and biomedical VLMs and 3 independent benchmarks: VQA-RAD~\cite{lau2018vqarad}, SLAKE~\cite{liu2021slake}, and VQA-Med-2019~\cite{benabacha2019vqamed}.

\begin{table*}[t]
\centering
\caption{Close-ended benchmark summary by method. AUROC is error-detection AUROC, AURC/E-AURC are selective-risk areas, and Cov@r is maximum coverage at selective risk $\leq r$.}
\label{tab:benchmark_main_summary}
\small
\begin{tabular}{llrrrrrrrr}
\toprule
Task & Method & Settings & Acc. & Base risk & AUROC $\uparrow$ & AURC $\downarrow$ & E-AURC $\downarrow$ & Cov@5\% $\uparrow$ & Cov@10\% $\uparrow$ \\
\midrule
Closed & Sampling-SE & 21 & 0.751 & 0.249 & 0.555 & 0.235 & 0.195 & 0.037 & 0.142 \\
Closed & PTrue & 21 & 0.751 & 0.249 & 0.616 & 0.171 & 0.132 & 0.205 & 0.312 \\
Closed & HAC & 21 & 0.751 & 0.249 & 0.556 & 0.240 & 0.200 & 0.023 & 0.121 \\
Closed & HAC-Platt & 21 & 0.751 & 0.249 & 0.564 & 0.227 & 0.187 & 0.030 & 0.134 \\
Closed & HAC-Gate & 21 & 0.751 & 0.249 & 0.575 & 0.222 & 0.183 & 0.036 & 0.179 \\
Closed & CAS & 21 & 0.751 & 0.249 & 0.736 & 0.131 & 0.092 & 0.319 & 0.496 \\
\bottomrule
\vspace{-5mm}
\end{tabular}
\end{table*}

\begin{table*}[t]
\centering
\caption{Closed-ended HAC versus CAS results with paired bootstrap confidence intervals. AUROC is error-detection AUROC; HAC best is selected within HAC/HAC-Platt/HAC-Gate under the same cross-validation protocol.}
\label{tab:closed_results}
\small
\begin{tabular}{llrrrrrr}
\toprule
Dataset & Model & $n$ & Acc. & HAC best & CAS & $\Delta$ & 95\% CI \\
\midrule
\multirow{7}{*}{VQA-RAD} & Qwen3-VL-2B  & 250 & 0.712 & 0.574 & 0.646 & +0.072 & [-0.023, +0.170] \\
 & Qwen3-VL-8B  & 249 & 0.747 & 0.548 & 0.689 & +0.141 & [+0.047, +0.230] \\
 & InternVL3-2B & 249 & 0.715 & 0.603 & 0.652 & +0.050 & [-0.030, +0.140] \\
 & InternVL3-8B & 249 & 0.787 & 0.565 & 0.717 & +0.153 & [+0.060, +0.247] \\
 & Lingshu-7B   & 250 & 0.780 & 0.535 & 0.680 & +0.146 & [+0.061, +0.234] \\
 & MedGemma-4B  & 250 & 0.804 & 0.630 & 0.662 & +0.032 & [-0.069, +0.131] \\
 & MedVLM-R1    & 250 & 0.628 & 0.511 & 0.630 & +0.119 & [+0.046, +0.199] \\
\midrule
\multirow{7}{*}{SLAKE} & Qwen3-VL-2B  & 344 & 0.811 & 0.658 & 0.775 & +0.117 & [+0.036, +0.202] \\
 & Qwen3-VL-8B  & 355 & 0.794 & 0.621 & 0.768 & +0.147 & [+0.065, +0.230] \\
 & InternVL3-2B & 355 & 0.758 & 0.669 & 0.777 & +0.108 & [+0.059, +0.160] \\
 & InternVL3-8B & 355 & 0.808 & 0.628 & 0.794 & +0.166 & [+0.087, +0.249] \\
 & Lingshu-7B   & 355 & 0.842 & 0.683 & 0.876 & +0.193 & [+0.117, +0.274] \\
 & MedGemma-4B  & 352 & 0.869 & 0.577 & 0.682 & +0.105 & [-0.007, +0.224] \\
 & MedVLM-R1    & 355 & 0.690 & 0.547 & 0.767 & +0.220 & [+0.141, +0.297] \\
\midrule
\multirow{7}{*}{VQA-Med-2019} & Qwen3-VL-2B  & 64  & 0.688 & 0.544 & 0.581 & +0.037 & [-0.130, +0.200] \\
 & Qwen3-VL-8B  & 64  & 0.734 & 0.659 & 0.845 & +0.186 & [-0.010, +0.402] \\
 & InternVL3-2B & 64  & 0.797 & 0.652 & 0.837 & +0.185 & [+0.056, +0.314] \\
 & InternVL3-8B & 64  & 0.875 & 0.564 & 0.917 & +0.354 & [+0.173, +0.525] \\
 & Lingshu-7B   & 64  & 0.766 & 0.490 & 0.807 & +0.317 & [+0.181, +0.436] \\
 & MedGemma-4B  & 64  & 0.594 & 0.514 & 0.623 & +0.109 & [-0.066, +0.285] \\
 & MedVLM-R1    & 64  & 0.578 & 0.537 & 0.728 & +0.191 & [+0.044, +0.314] \\
\midrule
\textbf{Macro mean} & -- & -- & -- & \textbf{0.586} & \textbf{0.736} & \textbf{+0.150} & -- \\
\bottomrule
\end{tabular}
\vspace{-5mm}
\end{table*}

\begin{figure*}
    \centering
    \includegraphics[width=1\linewidth]{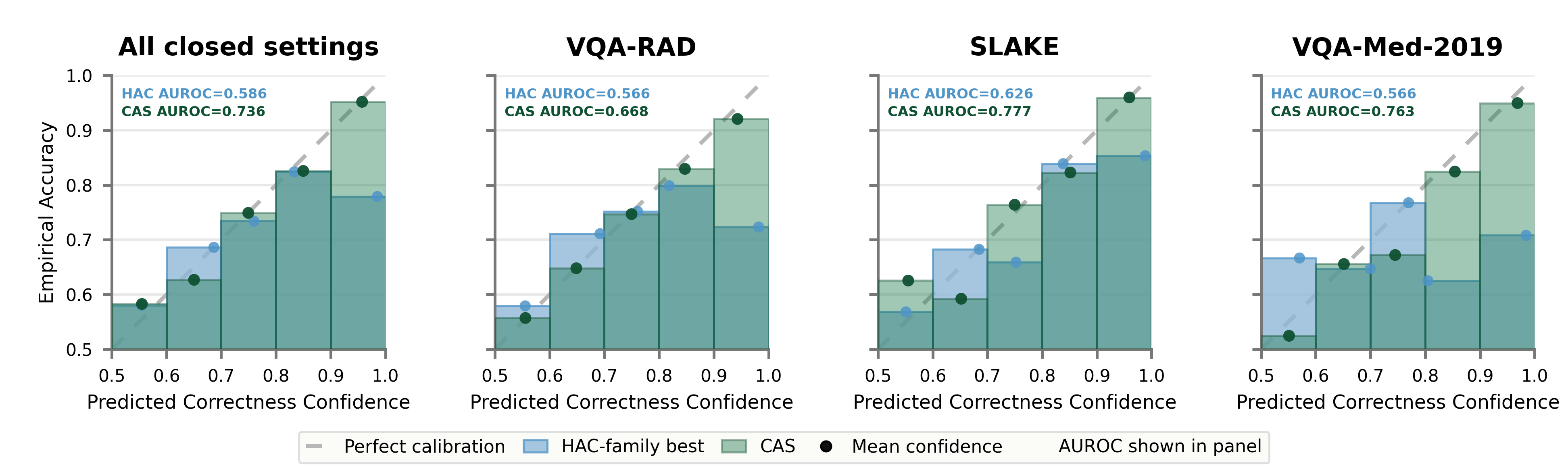}
    \caption{Calibration of predicted correctness across closed-ended benchmarks. Bars show empirical accuracy within predicted-correctness bins for HAC-family
  best and CAS scores; dots denote mean confidence in each bin; AUROC values are shown
  in the upper-left corner of each panel and follow the same 21-setting closed-ended benchmark protocol as Table~\ref{tab:closed_results}. We show the high-confidence regime ($0.5$--$1.0$), corresponding to the selective-
  prediction acceptance region. CAS improves error-detection AUROC over the best HAC-family score while also producing predicted-correctness estimates that better track empirical accuracy in
  high-confidence regions.}
    \label{fig:calibration_overlay}
\end{figure*}

Table~\ref{tab:benchmark_main_summary} reports the aggregate closed-ended results, and Table~\ref{tab:closed_results} provides the per-model, per dataset breakdown.  We also report selective-prediction metrics following the risk--coverage formulation~\cite{elyaniv2010foundations}. The central finding is that CAS consistently outperforms the best HAC-family score and standard calibration alternatives on all 21 model-dataset cells. On average, CAS achieves a 0.150 AUROC gain over the best-performing sampling-stability metric for clinical error detection. This aggregate gain is consistent with the stable-wrongness diagnosis: sampling-stability methods can hover close to random chance in closed-ended settings because stable answers are not necessarily grounded answers. Post-hoc monotonic scaling (HAC-Platt) yields virtually zero downstream ranking improvements~\cite{guo2017calibration,nixon2019measuring,byun2026overconfidence}, supporting our claim that adjusting scalar confidence values alone cannot resolve structurally stable ranking failures. The next section tests whether the gain comes from recovering HAC-confident hidden errors and from counterfactual boundary-shift signals.

Figure~\ref{fig:calibration_overlay} complements the ranking-based AUROC results with a calibration view.  CAS aligns predicted correctness with empirical
accuracy more closely than the HAC-family best score, particularly in the high-confidence region where selective prediction would accept model answers.

\section{Why CAS Works}
\subsection{Stable Wrongness Breaks Sampling-Stability Reliability}

The performance gap between HAC and CAS follows from the failure mode diagnosed in Section~\ref{sec:rethinking}. %Sampling-based reliability methods, including HAC and semantic-entropy-style estimators, are most effective when incorrect answers are unstable: repeated high-temperature generations expose disagreement, and disagreement becomes an uncertainty signal~\cite{wang2023selfconsistency,kuhn2023semantic,farquhar2024detecting,liao2025vase}. Closed-ended medical VQA creates a harder case. The answer space is compressed to two reusable labels, and the model can preserve a confident yes/no preference even when the paired image is removed, blacked out, or replaced by an unrelated image. 
In this regime, answer stability is ambiguous: it may indicate visual grounding, but it may also indicate stable wrongness.

CAS addresses this ambiguity by measuring how the support for the committed answer and its yes/no alternative changes when visual evidence is removed or corrupted. We investigate two questions: whether CAS recovers HAC-confident errors and which counterfactual signals contribute most to its performance.

\subsection{CAS Recovers HAC-Confident Hidden Errors}

To examine whether CAS detects errors overlooked by HAC, we divide samples into four groups using the median HAC and CAS risks within each model–dataset setting, then aggregate results across all 21 settings. Lower risk indicates greater estimated reliability.
\begin{table}[t]
\centering
\caption{CAS identifies hidden errors among samples trusted by HAC. The H-low/C-high cell has low HAC risk but substantially higher error rate, indicating a HAC-confident failure region recovered by CAS.}
\label{tab:hac_cas_quadrants}
\small
\setlength{\tabcolsep}{2.6pt}
\begin{tabular}{@{}lrrrrr@{}}
\toprule
Quadrant & $n$ & Err. & HAC & CAS & Opp. shift \\
\midrule
H-low / C-low & 1728 & 0.111 & 0.128 & 0.229 & -5.769 \\
H-low / C-high & 1123 & 0.318 & 0.121 & 0.603 & -2.177 \\
H-high / C-low & 609 & 0.123 & 0.266 & 0.269 & -5.782 \\
H-high / C-high & 1206 & 0.384 & 0.317 & 0.657 & -1.735 \\
\bottomrule
\end{tabular}

\end{table}

Table~\ref{tab:hac_cas_quadrants} shows that CAS identifies errors even among HAC-confident predictions. The H-low/C-high group has nearly the same HAC risk as H-low/C-low (0.121 vs. 0.128), but an almost threefold error rate (0.318 vs. 0.111). CAS distinguishes these groups with substantially different risk scores (0.603 vs. 0.229). This pattern persists among the lowest-risk 20\% of HAC predictions: incorrect and correct answers have similar HAC risks (0.091 vs. 0.095), whereas CAS assigns higher risks to incorrect answers (0.545 vs. 0.348). These results suggest that counterfactual sensitivity helps CAS identify errors missed by sampling-based confidence estimates.

\subsection{High CAS Risk Reflects Counterfactual Boundary Weakness}

We characterize the features associated with CAS risk by comparing the lowest- and highest-risk 20\% of samples within each setting. Instead, Table~\ref{tab:cas_risk_behavior} shows a consistent behavioral pattern.

\begin{table}[t]
\centering
\caption{High CAS risk corresponds to counterfactual boundary weakness rather than hard-answer instability. Low and high denote the lowest and highest 20\% CAS-risk samples within each model-dataset setting.}
\label{tab:cas_risk_behavior}
\small
\setlength{\tabcolsep}{2.8pt}
\begin{tabular}{@{}lrrrrr@{}}
\toprule
Subset & $n$ & Err. & CF support & Opp. shift & Same pred. \\
\midrule
Low 20\% & 936 & 0.061 & 0.597 & -7.585 & 0.613 \\
High 20\% & 936 & 0.439 & 0.521 & -1.093 & 0.510 \\
\bottomrule
\end{tabular}
\end{table}

Notably, high-risk samples exhibit lower same-prediction rates (0.510 vs. 0.613), indicating that CAS risk is not simply determined by hard-answer persistence. The largest contrast appears in the opposite-answer shift (-1.093 vs. -7.585), suggesting weaker counterfactual decision-boundary responses among high-risk predictions. Because these groups are defined by CAS scores, the feature differences characterize the detector’s learned behavior rather than independently establishing a causal mechanism. The following ablation analysis further evaluates the predictive contributions of individual signal families.

\subsection{Counterfactual Boundary Shift Drives CAS}

To identify the contributions of different counterfactual signals, we evaluate four feature families using the same five-fold out-of-fold logistic-regression protocol as the main CAS model: target-answer shift, opposite-answer shift, counterfactual support, and prediction stability. These respectively measure changes in committed-answer support, relative yes/no preference, residual support under corrupted inputs, and hard-answer persistence.

\begin{table}[t]
\centering
\caption{Signal-family ablation shows that CAS is driven by counterfactual boundary shift, not hard prediction stability. CAS-paper-9D is the default compact representation and CAS-full-17D adds raw condition-specific target and opposite-answer shifts. Metrics are macro-averaged over 7 models and 3 closed-ended datasets.}
\label{tab:cas_signal_ablation}
\small
\setlength{\tabcolsep}{3.0pt}
\begin{tabular}{@{}lrrrr@{}}
\toprule
Feature group & Dim. & AUROC $\uparrow$ & AURC $\downarrow$ & E-AURC $\downarrow$ \\
\midrule
Target-answer shift & 7 & 0.605 & 0.205 & 0.166 \\
Opposite-answer shift & 7 & 0.711 & 0.146 & 0.106 \\
Counterfactual support & 2 & 0.577 & 0.212 & 0.172 \\
Prediction stability & 1 & 0.538 & 0.243 & 0.204 \\
\midrule
CAS-paper-9D & 9 & 0.736 & 0.131 & 0.092 \\
CAS-full-17D & 17 & 0.741 & 0.131 & 0.091 \\
\bottomrule
\end{tabular}
\end{table}

Table~\ref{tab:cas_signal_ablation} shows that opposite-answer shift is the strongest standalone feature family, achieving 0.711 AUROC. Target-answer shift and direct counterfactual support achieve 0.605 and 0.577, respectively, while prediction stability performs near chance (0.538). These results indicate that changes in the relative yes/no decision boundary are more informative than either target-support changes alone or binary prediction stability.

Combining these signals, the compact 9-dimensional CAS representation achieves 0.736 AUROC, compared with 0.741 for the full 17-dimensional variant. The marginal improvement from additional condition-specific features supports the compact representation and suggests that CAS gains arise primarily from informative counterfactual signals rather than feature dimensionality alone.

\subsection{Limitations}
CAS is strongest when the answer space has a natural contrast. This is why the current method claim is restricted to closed-ended yes/no VQA. 

Counterfactual corruptions are also not perfect causal interventions. A black image, random image, or blurred image may introduce its own input-state cue. We treat these probes as practical diagnostics of evidence sensitivity, not as proof of a complete causal mechanism.

\section{Conclusion}

In this study, we presented an empirical audit of closed-ended medical VQA evaluation and exposed stable wrongness as a structural reliability failure across modern VLM families. Through counterfactual-image diagnostics, we showed that binary yes/no answers can remain confident and repeatable even when patient-specific visual evidence is removed or mismatched. This creates a failure mode for uncertainty frameworks such as HAC: when wrong answers are stable, sampling agreement can remain deceptively elevated, turning answer stability itself into an unreliable signal. To address this vulnerability, we introduced Counterfactual Answer Sensitivity (CAS), a metric that shifts reliability estimation from answer stability to evidence sensitivity. By mapping target-support and opposite-answer boundary shifts across a parallel intervention space, including text-only, black, noisy, and blurred visual states, CAS systematically flags predictions whose support is weakly tied to the real image. Across 7 VLMs and 3 datasets, CAS outperformed traditional likelihood and sampling baselines, yielding a macro-averaged AUROC gain of 0.150 for clinical error detection. The gain is concentrated in HAC-confident hidden errors and is driven mainly by counterfactual decision-boundary shift rather than hard prediction stability.


\begin{thebibliography}{00}

\bibitem{alayrac2022flamingo}
J.-B. Alayrac, J.~Donahue, P.~Luc, A.~Miech, I.~Barr, Y.~Hasson, K.~Lenc, A.~Mensch, K.~Millican, M.~Reynolds, R.~Ring, E.~Rutherford, S.~Cabi, T.~Han, Z.~Gong, S.~Samangooei, M.~Monteiro, J.~Menick, S.~Borgeaud, A.~Brock, A.~Nematzadeh, S.~Sharifzadeh, M.~Binkowski, R.~Barreira, O.~Vinyals, A.~Zisserman, and K.~Simonyan, ``Flamingo: A visual language model for few-shot learning,'' in \emph{Advances in Neural Information Processing Systems}, vol.~35, 2022, pp. 23716--23736.
\bibitem{li2023blip2}
J.~Li, D.~Li, S.~Savarese, and S.~C.~H. Hoi, ``BLIP-2: Bootstrapping language-image pre-training with frozen image encoders and large language models,'' in \emph{International Conference on Machine Learning}, 2023, pp. 19730--19742.
\bibitem{liu2023llava}
H.~Liu, C.~Li, Q.~Wu, and Y.~J. Lee, ``Visual instruction tuning,'' in \emph{Advances in Neural Information Processing Systems}, vol.~36, 2023, pp. 34892--34916.
\bibitem{li2023llavamed}
C.~Li, C.~Wong, S.~Zhang, N.~Usuyama, H.~Liu, J.~Yang, T.~Naumann, H.~Poon, and J.~Gao, ``LLaVA-Med: Training a large language-and-vision assistant for biomedicine in one day,'' in \emph{Advances in Neural Information Processing Systems}, vol.~36, 2023, pp. 28541--28564.
\bibitem{tu2024medpalm}
T.~Tu \emph{et al.}, ``Towards generalist biomedical AI,'' \emph{NEJM AI}, vol.~1, no.~3, Art.~no.~AIoa2300138, 2024, doi: 10.1056/AIoa2300138.
\bibitem{zhu2025trust}
Z.~Zhu, Y.~Zhang, X.~Zhuang, F.~Zhang, Z.~Wan, Y.~Chen, Q.~Long, Y.~Zheng, and X.~Wu, ``Can we trust AI doctors? A survey of medical hallucination in large language and large vision-language models,'' in \emph{Findings of the Association for Computational Linguistics: ACL 2025}, 2025, pp. 6748--6769, doi: 10.18653/v1/2025.findings-acl.350.
\bibitem{byun2026overconfidence}
J.~Y. Byun, Y.-J. Park, J.-P. Corbeil, and A.~Ben Abacha, ``Overconfidence and calibration in medical VQA: Empirical findings and hallucination-aware mitigation,'' \emph{arXiv preprint arXiv:2604.02543}, 2026.
\bibitem{agrawal2018dont}
A.~Agrawal, D.~Batra, D.~Parikh, and A.~Kembhavi, ``Don't just assume; look and answer: Overcoming priors for visual question answering,'' in \emph{Proceedings of the IEEE/CVF Conference on Computer Vision and Pattern Recognition}, 2018, pp. 4971--4980.
\bibitem{goyal2017making}
Y.~Goyal, T.~Khot, D.~Summers-Stay, D.~Batra, and D.~Parikh, ``Making the V in VQA matter: Elevating the role of image understanding in visual question answering,'' in \emph{Proceedings of the IEEE/CVF Conference on Computer Vision and Pattern Recognition}, 2017, pp. 6904--6913.
\bibitem{cadene2019rubi}
R.~Cadene, C.~Dancette, H.~Ben-younes, M.~Cord, and D.~Parikh, ``RUBi: Reducing unimodal biases in visual question answering,'' in \emph{Advances in Neural Information Processing Systems}, vol.~32, 2019.
\bibitem{geirhos2020shortcut}
R.~Geirhos, J.-H. Jacobsen, C.~Michaelis, R.~Zemel, W.~Brendel, M.~Bethge, and F.~A. Wichmann, ``Shortcut learning in deep neural networks,'' \emph{Nature Machine Intelligence}, vol.~2, no.~11, pp. 665--673, 2020, doi: 10.1038/s42256-020-00257-z.
\bibitem{zech2018variable}
J.~R. Zech, M.~A. Badgeley, M.~Liu, A.~B. Costa, J.~J. Titano, and E.~K. Oermann, ``Variable generalization performance of a deep learning model to detect pneumonia in chest radiographs: A cross-sectional study,'' \emph{PLOS Medicine}, vol.~15, no.~11, article e1002683, 2018, doi: 10.1371/journal.pmed.1002683.
\bibitem{degrave2021ai}
A.~J. DeGrave, J.~D. Janizek, and S.-I. Lee, ``AI for radiographic COVID-19 detection selects shortcuts over signal,'' \emph{Nature Machine Intelligence}, vol.~3, pp. 610--619, 2021, doi: 10.1038/s42256-021-00338-7.
\bibitem{ross2017right}
A.~S. Ross, M.~C. Hughes, and F.~Doshi-Velez, ``Right for the right reasons: Training differentiable models by constraining their explanations,'' in \emph{International Joint Conference on Artificial Intelligence}, 2017, pp. 2662--2670.
\bibitem{kuhn2023semantic}
L.~Kuhn, Y.~Gal, and S.~Farquhar, ``Semantic uncertainty: Linguistic invariances for uncertainty estimation in natural language generation,'' in \emph{International Conference on Learning Representations}, 2023.
\bibitem{farquhar2024detecting}
S.~Farquhar, J.~Kossen, L.~Kuhn, and Y.~Gal, ``Detecting hallucinations in large language models using semantic entropy,'' \emph{Nature}, vol.~630, pp. 625--630, 2024, doi: 10.1038/s41586-024-07421-0.
\bibitem{liao2025vase}
Z.~Liao, S.~Hu, K.~Zou, H.~Fu, L.~Zhen, and Y.~Xia, ``Vision-amplified semantic entropy for hallucination detection in medical visual question answering,'' in \emph{Medical Image Computing and Computer Assisted Intervention -- MICCAI 2025}, LNCS 15964, pp. 669--679, Springer Nature Switzerland, 2025, doi: 10.1007/978-3-032-04971-1\_63.
\bibitem{guo2017calibration}
C.~Guo, G.~Pleiss, Y.~Sun, and K.~Q. Weinberger, ``On calibration of modern neural networks,'' in \emph{International Conference on Machine Learning}, 2017, pp. 1321--1330.
\bibitem{nixon2019measuring}
J.~Nixon, M.~W. Dusenberry, L.~Zhang, G.~Jerfel, and D.~Tran, ``Measuring calibration in deep learning,'' in \emph{Proceedings of the IEEE/CVF Conference on Computer Vision and Pattern Recognition Workshops}, 2019, pp. 38--41.
\bibitem{antol2015vqa}
S.~Antol, A.~Agrawal, J.~Lu, M.~Mitchell, D.~Batra, C.~L. Zitnick, and D.~Parikh, ``VQA: Visual question answering,'' in \emph{Proceedings of the IEEE International Conference on Computer Vision}, 2015, pp. 2425--2433.
\bibitem{lau2018vqarad}
J.~J. Lau, S.~Gayen, A.~Ben Abacha, and D.~Demner-Fushman, ``A dataset of clinically generated visual questions and answers about radiology images,'' \emph{Scientific Data}, vol.~5, article 180251, 2018, doi: 10.1038/sdata.2018.251.
\bibitem{liu2021slake}
B.~Liu, L.-M. Zhan, L.~Xu, L.~Ma, Y.~Yang, and X.-M. Wu, ``SLAKE: A semantically-labeled knowledge-enhanced dataset for medical visual question answering,'' in \emph{IEEE International Symposium on Biomedical Imaging}, 2021, pp. 1650--1654.
\bibitem{benabacha2019vqamed}
A.~Ben Abacha, S.~A. Hasan, V.~V. Datla, J.~Liu, D.~Demner-Fushman, and H.~M{\"u}ller, ``VQA-Med: Overview of the medical visual question answering task at ImageCLEF 2019,'' in \emph{CLEF 2019 Working Notes}, CEUR Workshop Proceedings, vol.~2380, 2019.
\bibitem{hendrycks2017baseline}
D.~Hendrycks and K.~Gimpel, ``A baseline for detecting misclassified and out-of-distribution examples in neural networks,'' in \emph{International Conference on Learning Representations}, 2017.
\bibitem{ovadia2019uncertainty}
Y.~Ovadia, E.~Fertig, J.~Ren, Z.~Nado, D.~Sculley, S.~Nowozin, J.~V. Dillon, B.~Lakshminarayanan, and J.~Snoek, ``Can you trust your model's uncertainty? Evaluating predictive uncertainty under dataset shift,'' in \emph{Advances in Neural Information Processing Systems}, vol.~32, 2019.
\bibitem{wang2023selfconsistency}
X.~Wang, J.~Wei, D.~Schuurmans, Q.~Le, E.~Chi, S.~Narang, A.~Chowdhery, and D.~Zhou, ``Self-consistency improves chain of thought reasoning in language models,'' in \emph{International Conference on Learning Representations}, 2023.
\bibitem{huang2026euq}
T.~Huang, R.~Wang, X.~Liu, Y.~Qin, L.~Duan, and L.~Jing, ``Detecting misbehaviors of large vision-language models by evidential uncertainty quantification,'' in \emph{International Conference on Learning Representations}, 2026, arXiv:2602.05535.
\bibitem{goyal2019counterfactual}
Y.~Goyal, Z.~Wu, J.~Ernst, D.~Batra, D.~Parikh, and S.~Lee, ``Counterfactual visual explanations,'' in \emph{International Conference on Machine Learning}, 2019, pp. 2376--2384.
\bibitem{elyaniv2010foundations}
R.~El-Yaniv and Y.~Wiener, ``On the foundations of noise-free selective classification,'' \emph{Journal of Machine Learning Research}, vol.~11, pp. 1605--1641, 2010.

\end{thebibliography}
\end{document}